\documentclass[10pt,twocolumn,letterpaper]{article}

\usepackage[pagenumbers]{cvpr} 

\usepackage[dvipsnames]{xcolor}

\newcommand{\myparagraph}[1]{\noindent{\bf #1}}

\definecolor{lightgray}{gray}{0.92}

\usepackage{multirow}
\usepackage{xcolor}
\usepackage{algorithm}
\usepackage{algorithmic}
\usepackage{lineno}
\usepackage[dvipsnames]{xcolor}
\usepackage{amsmath}
\usepackage{colortbl}
\usepackage{natbib}

\newcommand\blfootnote[1]{%
\begingroup
\renewcommand\thefootnote{}\footnote{#1}%
\addtocounter{footnote}{-1}%
\endgroup
}
\usepackage{amssymb}  
\usepackage{bbding} 
\usepackage{pifont} 
\definecolor{cvprblue}{rgb}{0.21,0.49,0.74}
\usepackage[pagebackref,breaklinks,colorlinks,citecolor=cvprblue]{hyperref}

\title{An Efficient Occupancy World Model via Decoupled Dynamic Flow and Image-assisted Training}

\author{
{Haiming Zhang}$^{1,2*}$,
{Ying Xue}$^{1,2}$,
{Xu Yan}$^{4\textsuperscript{\ding{41}}}$,
{Jiacheng Zhang}$^{3}$\\
{Weichao Qiu}$^{4}$,
{Dongfeng Bai}$^{4}$,
{Bingbing Liu}$^{4}$,
{Shuguang Cui}$^{2,1}$,
{Zhen Li}$^{2,1\textsuperscript{\ding{41}}}$\\
$^1$FNii, Shenzhen~~
$^2$SSE, CUHK-Shenzhen\\
$^3$HKU~~
$^4$Huawei Noah’s Ark Lab\\
}

\begin{document}
\maketitle

\begin{abstract}
The field of autonomous driving is experiencing a surge of interest in world models, which aim to predict potential future scenarios based on historical observations. 
In this paper, we introduce \textbf{DFIT-OccWorld}, an efficient 3D \textbf{Occ}upancy \textbf{World} model that leverages \textbf{D}ecoupled dynamic \textbf{F}low and \textbf{I}mage-assisted \textbf{T}raining strategy, substantially improving 4D scene forecasting performance. 
To simplify the training process, we discard the previous two-stage training strategy and innovatively reformulate the occupancy forecasting problem as a decoupled voxels warping process.
Our model forecasts future dynamic voxels by warping existing observations using voxel flow, whereas static voxels are easily obtained through pose transformation.
Moreover, our method incorporates an image-assisted training paradigm to enhance prediction reliability. 
Specifically, differentiable volume rendering is adopted to generate rendered depth maps through predicted future volumes, which are adopted in render-based photometric consistency.
Experiments demonstrate the effectiveness of our approach, showcasing its state-of-the-art performance on the nuScenes and OpenScene benchmarks for 4D occupancy forecasting, end-to-end motion planning and point cloud forecasting. 
Concretely, it achieves state-of-the-art performances compared to existing 3D world models while incurring substantially lower computational costs.
\blfootnote{$^*$Work done during an internship at Huawei Noah’s Ark Lab.}
\blfootnote{$\textsuperscript{\ding{41}}$Corresponding authors.}
\end{abstract}

\section{Introduction}
\label{sec:intro}

\begin{figure}[t]
    \centering
    \includegraphics[width=0.9\linewidth]{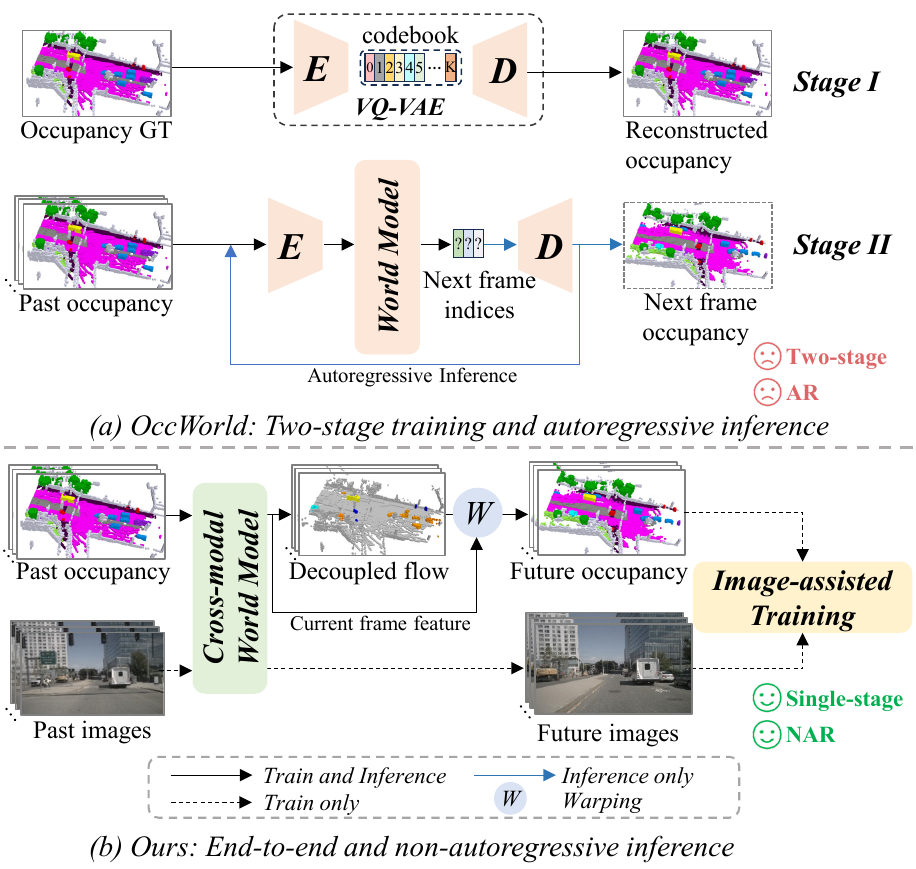}
    \vspace{-0.2cm}
    \caption{\textbf{Comparison with existing methods.} (a) The pipeline of the existing method, \ie, OccWorld, adopts a two-stage training paradigm and autoregressive (AR) manner for inference. \textit{E} and \textit{D} denote the encoder, and decoder, respectively. (b) Our method utilizes an end-to-end and non-autoregressive (NAR) pipeline with the decoupled dynamic flow strategy for occupancy forecasting. The proposed image-assisted training enhances the performance.}
    \label{fig:teaser}
    \vspace{-0.25cm}
\end{figure}

Autonomous driving has the potential to revolutionize transportation by addressing numerous real-world traffic problems~\cite{dong2023applications}. 
To effectively navigate the complex and dynamic environment, autonomous vehicles require the capability to understand the current surrounding environment and forecast how it will evolve to the future. 
This critical competency has spurred the recent emergence of generative world models. 
While initially proposed for simulated control or real-world robotics applications~\cite{ha2018world,wu2023daydreamer}, world models have seen significant advancements in the autonomous driving field and have been extended to a more general concept: forecast future scenarios based on past observations~\cite{yan2024forging,guan2024world}. 

Most of the early world models focused on future video prediction~\cite{hu2023gaia,wang2023drivedreamer,wang2024driving}, ignoring the crucial 3D information necessary for autonomous driving.
To alleviate this, some approaches~\cite{khurana2023point,agro2024uno} are dedicated to predicting future point clouds by constructing a 4D volume feature from past point cloud inputs. 
Further, ViDAR~\cite{yang2023vidar} formulates a novel visual point cloud forecasting task that predicts future point clouds from historical visual images, enabling scalable autonomous driving.
Although involving 3D information, they forecast futures in an implicit manner, which lacks interpretability and efficiency. And predicting solely geometric point clouds devoid of semantic context.

3D semantic occupancy, encoding both the occupancy state and semantic information within a 3D volume, emerges as a compelling representation for describing 3D scenes of autonomous driving. 
Building upon this concept, Zheng \etal proposed OccWorld~\cite{zheng2023occworld}, a 3D world model that simultaneously predicts future occupancy and plans the trajectory of the ego vehicle based on past occupancy observations.
Specifically, OccWorld employs a two-stage training strategy, as shown in Fig.~\ref{fig:teaser}(a):
In the first stage, a vector-quantized variational autoencoder (VQ-VAE)~\cite{van2017neural} acts as an occupancy tokenizer, learning discrete scene tokens in a self-supervised manner. This obtains high-level representations from the observed occupancy inputs.
In the second stage, the 4D occupancy forecasting problem is recast as a classification task predicting codebook indices, where a spatial-temporal generative transformer is utilized as a world model to forecast future occupancy.
While showcasing promising results, OccWorld faces limitations impacting its real-world flexibility and scalability:
\begin{itemize}
    \setlength{\itemsep}{0pt}
    \setlength{\parsep}{0pt}
    \setlength{\parskip}{0pt}
    \item The VQ-VAE training in the first stage significantly hinders efficiency and introduces performance bottlenecks for the second stage.  Finding suitable hyperparameters for the scene tokenizer remains challenging in balancing reconstruction and forecasting performance~\cite{zheng2023occworld}.
    \item OccWorld relies solely on implicit features in an autoregressive manner. It either ignores the explicit structural consistency in adjacent scenes or texture information readily available from images. These hinder the model's ability to fully capture the dynamics of the environment.
\end{itemize}

To address the shortcomings identified in OccWorld, we introduce a novel end-to-end 3D occupancy world model in autonomous driving. 
This framework builds upon the advantages and potential of single-stage video prediction, enabling the simultaneous prediction of multiple future volumes and images, as shown in Fig.~\ref{fig:teaser}(b). This design facilitates information sharing among future predictions, enhancing both the accuracy and rationality of the results.
Moreover, we found that directly predicting the occupancy of each frame results in unsatisfactory performance due to the majority of the voxels being
empty. To address this issue, we use the semantic information predicted by the occupancy network to decouple voxels into dynamic and static categories. The world model then only predicts the voxel flow of dynamic objects and warps these voxels accordingly. For static objects, since their global positions remain unchanged, we can easily obtain them through pose transformation.
Furthermore, we propose an image-assisted training strategy that incorporates images as auxiliary input during the training phase.
Specifically, we employ differentiable volume rendering to obtain depth maps from forecasted future volumes.
Building on this, rendering-based photometric consistency is proposed to enhance the temporal consistency across multiple occupancy predictions. This process projects adjacent source frames onto target frames based on the rendered depths and calculates the reconstruction error between the re-projected images and the target images.
By leveraging the above components, our method exceeds existing methods by a large margin. 

The main contributions of this paper are as follows:
\begin{itemize}
    \setlength{\itemsep}{0pt}
    \setlength{\parsep}{0pt}
    \setlength{\parskip}{0pt}
    \item We introduce an end-to-end 3D occupancy world model specifically designed for 4D scene forecasting, offering both efficiency and boosted performance.
    \item Building upon this foundation, the proposed decoupled dynamic flow strategy simplifies the training process and enhances prediction reliability. Meanwhile, we propose an image-assisted learning strategy that leverages images as auxiliary inputs during training.
    \item Our comprehensive approach achieves state-of-the-art performance on the nuScenes and OpenScene benchmarks for three tasks, surpassing existing world models by a significant margin. 
\end{itemize}

\begin{figure*}[tbp]
	\begin{center}
		\includegraphics[width=0.9\linewidth]{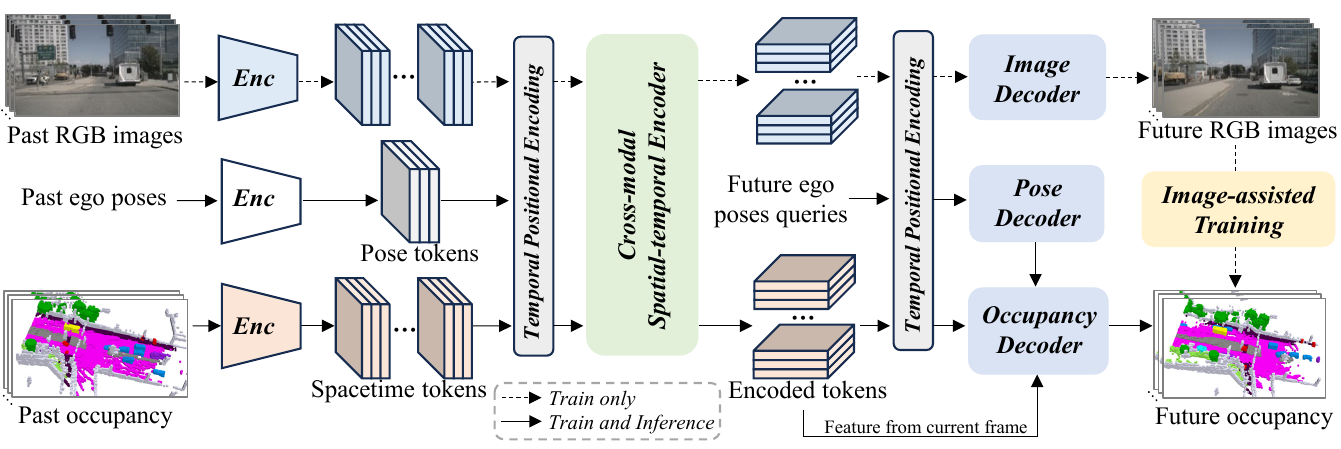}
	\end{center}
	\vspace{-.45cm}
	\caption{\textbf{Overall pipeline of our method.} It takes past images, occupancy, and ego poses as inputs, learning cross-modal information through a cross-modal spatial-temporal encoder. Then different decoders are responsible for predicting future frames in a non-autoregressive manner. And the proposed flow decoder and warping operation contained in the occupancy decoder could facilitate future occupancy forecasting. Additionally, an image-assisted training paradigm effectively imposes constraints on forecasted occupancy from the image domain, making more reliable occupancy predictions.}
	\label{fig2:framework}
\end{figure*}

\section{Related Work}
\label{sec:related}
\subsection{3D Semantic Occupancy Prediction}
The objective of 3D occupancy prediction is to ascertain the status of each voxel within a 3D space, specifically determining its occupancy and, if applicable, assigning semantic labels to occupied voxels. It has gained much popularity in recent years due to its ability to represent objects of any shape~\cite{shi2024effocc,tian2023occ3d,wang2023openoccupancy,tong2023scene}, especially for the vision-based occupancy prediction methods~\cite{zhang2024vision,cao2022monoscene,zhang2023occformer,zhang2023radocc,wang2023panoocc}, owing to their cost-effectiveness and adaptability. 
Previous works utilize dense volume representations obtained by multi-scale encoders~\cite{wei2023surroundocc}, or a novel tri-perspective view features interpolation~\cite{huang2023tri}, or a hybrid of forward and backward view transformation for occupancy prediction.
Afterwards, more and more novel approaches have been proposed. For example, SparseOcc~\cite{liu2023fully} exploits the fully sparse occupancy network. OSP~\cite{shi2024occupancysetpoints} presents the points of interest to represent the scene.
GaussianFormer~\cite{huang2024gaussian} proposes an object-centric representation to describe 3D scenes with sparse 3D semantic Gaussians.
More recently, several approaches~\cite{liu2024let,gan2023simple,pan2023renderocc,zhang2023occnerf,huang2023selfocc} have explored volume rendering technique~\cite{mildenhall2021nerf} for occupancy prediction in a self-supervised manner.
Existing methods, while impressive, fail to account for the temporal evolution of the scene. In this paper, we introduce an innovative model that not only facilitates the forecasting of future 4D occupancy and point clouds but also supports end-to-end autonomous driving.

\subsection{World Models in Autonomous Driving}
The world models, long-established in the robot field, aim to learn a compact representation of the world and predict its future states based on the agent's actions and past observations~\cite{ha2018world,hafner2019dream,wu2023daydreamer}.
Recently, learning world models in driving scenes have gained attention, and can be broadly categorized into 2D-based, 3D-based, and multi-modal approaches.
As a pioneering 2D image-based world model, GAIA-1~\cite{hu2023gaia} employs the video diffusion model to forecast future driving videos by taking video, text, and action as inputs.
Subsequent works like ADriver-I~\cite{jia2023adriver}, DriveDreamer~\cite{wang2023drivedreamer}, and Drive-WM~\cite{wang2024driving} improve prediction performance by incorporating control signal prediction or multi-view image forecasting. 
Meanwhile, research in 3D-based models is ongoing. Some methods predict future point clouds using discrete diffusion~\cite{zhang2023learning} or 4D volume representation~\cite{khurana2023point,agro2024uno} or latent rendering~\cite{yang2023vidar}.
While others explore 4D occupancy forecasting. For example, OccWorld~\cite{zheng2023occworld} employs a two-stage VQ-VAE framework for 3D semantic occupancy forecasting. LOPR~\cite{lange2024self} predicts future geometric occupancy grid maps. DriveWorld\cite{min2024driveworld} proposes a pre-training method via a complex 4D occupancy forecasting based on a complex state-space model.
Recently, multi-modal world models gained attention. MUVO~\cite{bogdoll2023muvo} integrates LiDAR point clouds beyond videos to predict future driving scenes in the representation of images, point clouds, and 3D occupancy. BEVWorld~\cite{zhang2024bevworld} unifies the multi-view images and LiDAR point clouds in the BEV representation for future images and point cloud forecasting.
The \textbf{concurrent} works \eg, RenderWorld~\cite{yan2024renderworld} and OccLLaMA~\cite{wei2024occllama} enhance occupancy forecasting performance through specialized encoder architectures. However, these approaches still adhere to the conventional two-stage training paradigm.

\section{Methodology}
\label{sec:method}

\subsection{Problem Formulation}
\label{sec:formulation}
In OccWorld~\cite{zheng2023occworld}, the world model $\mathcal{W}$ takes the historical sequence with $N_h$ 3D semantic occupancy $\mathcal{O}_{T-N_h:T}$ as inputs, along with the past ego trajectory sequence $\mathcal{P}_{T-N_h:T}$, to forecast the future occupancy and ego positions.
It adopts a two-stage training scheme and predicts future frames in an autoregressive manner as shown in Fig.~\ref{fig:teaser}(a). This process is formulated as:
\begin{align}
	\text{Stage I:} ~~~~~~~~~D(\mathbf{z}_T), \mathcal{C}  &= D(\mathbf{q}(E(\mathcal{O}_T))), \\
	\text{Stage II:} ~~~\mathbf{\hat{z}}_{T+1}, \mathcal{\hat{P}}_{T+1}  &= \mathcal{W}(\mathbf{z}_{T-N_h:T}, \mathcal{P}_{T-N_h:T}), \\
	\mathcal{\hat{O}}_{T+1} &= D(\mathbf{\hat{z}}_{T+1}, \mathcal{C} ),
	\label{eq:world_model}
\end{align}
where $E(\cdot)$ and $D(\cdot)$ represent encoder and decoder in VQ-VAE~\cite{van2017neural}. $\mathbf{q}(\cdot)$ is quantization operation, $\mathbf{z}_t$ is an approximate feature indices from a discrete codebook $\mathcal{C}$.
This framework is not flexible and scalable as the two-stage training strategy, and the autoregressive predictive strategy is susceptible to the error accumulation problem.

In this paper, we propose an end-to-end 3D occupancy world model with a single-stage training paradigm to model the evolution of the surrounding scenes. The above process can be simplified as:
%
\begin{multline}
    \mathcal{\hat{O}}_{T+1:T+{N_f}}, \mathcal{\hat{P}}_{T+1:T+N_f}  = \\
    \mathcal{W}^*(\mathcal{O}_{T-N_h:T}, \mathcal{P}_{T-N_h:T}).
    \label{eq:world_model2}
\end{multline}
However, it's inferior to regress future occupancy directly while ignoring the evidence that the majority of the voxels are empty. Therefore, our world model can be formulated as $\mathcal{W}^* = \mathcal{W}_p(\mathcal{W}_f)$, where $\mathcal{W}_f$ predicts the future flow maps with respect to the current frame, and the future occupancy can be obtained through warping operation $\mathcal{W}_p$:
\begin{align}
    \mathcal{\hat{F}}_{T+1:T+{N_f}}, \mathcal{\hat{P}}_{T+1:T+N_f}  &=
    \mathcal{W}_f(\mathcal{O}_{T-N_h:T}, \mathcal{P}_{T-N_h:T}), \\
    \mathcal{\hat{O}}_{T+1:T+{N_f}} &=  
    \mathcal{W}_p(\mathcal{O}_{T}, \mathcal{\hat{F}}_{T+1:T+{N_f}}).
    \label{eq:dd_flow}
\end{align}

Moreover, considering images are free and available in autonomous driving for occupancy prediction, we employ them as auxiliary inputs during the training process. Therefore, the training processes can be formulated as:
\begin{multline}
    \mathcal{\hat{O}}_{T+1:T+{N_f}}, \mathcal{\hat{P}}_{T+1:T+N_f}, \mathcal{\hat{I}}_{T+1:T+N_f}  = \\
    \mathcal{W}^*(\mathcal{O}_{T-N_h:T}, \mathcal{P}_{T-N_h:T}, \mathcal{I}_{T-N_h:T}),
    \label{eq:world_model1}
\end{multline}
where $\mathcal{I}_{T-N_h:T}$ is a historical image sequence.

\subsection{Framework Overview}
\label{sec:overview}
As depicted in Fig.~\ref{fig2:framework}, our method first preprocess historical 3D occupancy, image sequence, and ego poses inputs into spacetime tokens.
Then they are fed to a cross-modal spatial-temporal encoder to effectively capture spatial structures and local temporal dependencies among modalities.
After that, various decoders are leveraged to concurrently forecast future frames. 
To facilitate the learning process, we implement a decoupled dynamic flow strategy.
Moreover, an image-assisted training strategy is proposed to enhance occupancy forecasting without incurring computational burden during inference.

\subsection{Tokenization}
\label{sec:encoding}
\noindent\textbf{3D Occupancy Spacetime Encoding.}
Given a sequence of historically observed $N_h$ frames, 3D occupancy of each occupancy frame is denoted as $\mathcal{O}_i \in \mathbb{R}^{H_0 \times W_0 \times D_0}$, where $H_0$, $W_0$ and $D_0$ represent the resolution of the surrounding space centered on the ego car. Each voxel is assigned as one of $\mathcal{C}$ classes.
To encode the occupancy sequence into spacetime tokens, we first use a learnable class embedding to map the 3D occupancy into occupancy embedding $\hat{\mathbf{y}} \in \mathbb{R}^{H_0 \times W_0 \times D_0 \times C_0}$, where $C_0$ is the number of embedding channels.
Then, to reduce the computational burden, we transform the 3D occupancy embedding into the BEV representation $\mathbf{\tilde{y}} \in \mathbb{R}^{H_0 \times W_0 \times D_0C_0}$ following previous work~\cite{zheng2023occworld}. 
After that, the BEV embedding is decomposed into non-overlapping 2D patches $\mathbf{y}_p \in \mathbb{R}^{H \times W \times C^{\prime}}$, where $H=H_0/P$, $W=W_0/P$, $C^{\prime}=P^2\cdot D_0C_0$, and $P$ is the resolution of each image patch.
A lightweight encoder composed of several 2D convolution layers, \ie, Conv2d-GroupNorm-SiLU, is followed to extract the patch embeddings, obtaining the encoded historical occupancy spacetime tokens $\mathbf{{y}} \in \mathbb{R}^{N_h \times H \times W \times C}$.

\noindent\textbf{Image Spacetime Encoding.}
Given the same sequence length of historical images $\mathcal{I}_{T-N_h:T}$, we also process the 2D image modality into image spacetime tokens $\mathbf{x} \in \mathbb{R}^{N_h \times H \times W \times C}$.
Concretely, we resize the input images to the same size with BEV representation and decompose the 2D images into 2D patches.
Finally, we utilize an independent 2D encoder to extract the image patch embeddings for each frame.

\noindent\textbf{Ego Pose Encoding.}
In line with~\cite{zheng2023occworld}, we represent the ego pose as relative displacements between adjacent frames in the 2D ground plane.
Given historical ego poses, we utilize multiple linear layers followed by a ReLU activation function to obtain ego tokens $\mathbf{e} \in \mathbb{R}^{N_h \times C}$.

\begin{figure*}[tbp]
    \begin{center}
        \includegraphics[width=0.9\linewidth]{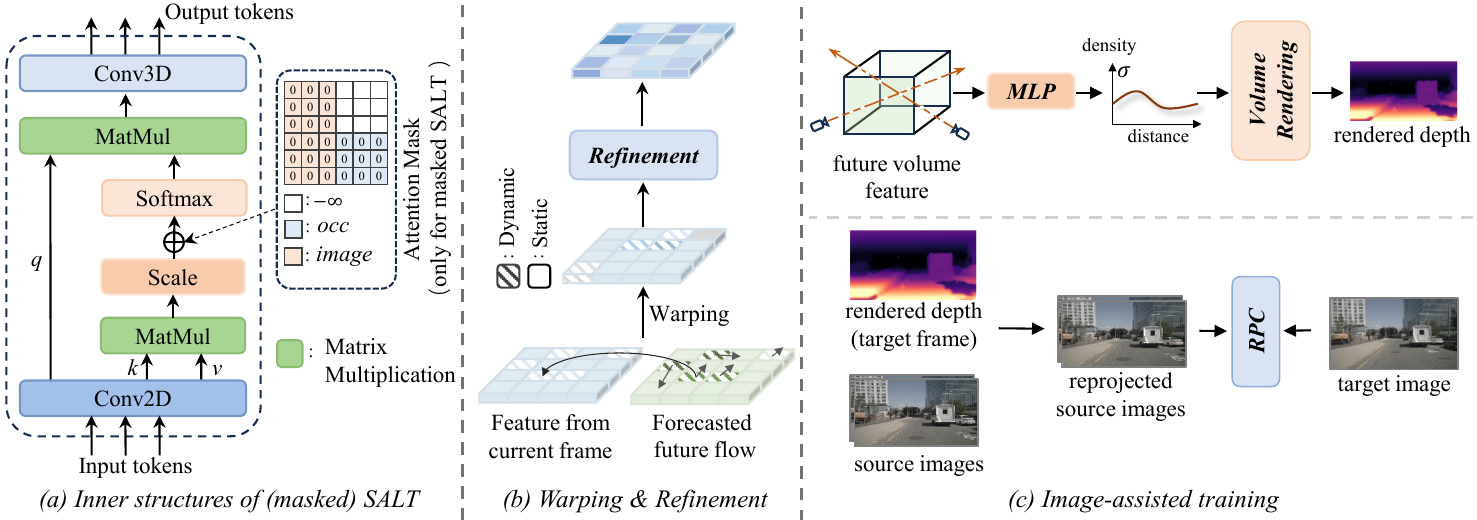}
    \end{center}
    \vspace{-0.25cm}
    \caption{\textbf{Inner structure of SALT, warping and refinement module, and the image-assisted training.} (a) The detailed structures of SALT, which replace the MLP and FFN in vanilla transformer with 2D convolutions and 3D convolutions respectively for capturing spatial-temporal dependencies. (b) We decouple the flow with the dynamic and static flow and warp the feature of the current frame for forecasting the future frame. The refinement module refines the coarse warping features. (c) The details of image-assisted training, where the Rendering-based Photometric Consistency (RPC) module is leveraged to further improve the forecasting performances with the depth map obtained by volume rendering. }
    \label{fig3:inner}
    \vspace{-0.25cm}
\end{figure*}

\subsection{Cross-Modal Spatial-Temporal Encoder}
\label{sec:encoder}

The cross-modal spatial-temporal encoder is responsible for not only capturing the spatial structures and temporal dependencies of inputs but also learning the correlation between different modalities. 

\noindent\textbf{Spatial-aware Local-temporal Attention Block.}
Inspired by previous works for video prediction~\cite{ning2022mimo}, we adopt a spatial-aware local-temporal (SALT) attention block in our encoder and decoder, as shown in Fig.~\ref{fig3:inner}(a). It first utilizes 2D convolution layers to generate the query map and paired key value for spacetime tokens. This spatial-aware CNN operation well retains the structural information.
Then, the standard multi-head attention is used to capture the temporal correlations between tokens.
Through such a manner, it learns the temporal correlation and preserves the spatial information of the sequence.
Moreover, it replaces the FFN (Feed Forward Network) layer with the 3DCNN to introduce local temporal clues for sequential modeling.

\myparagraph{Temporal Positional Encoding.}
We initialize a global embedding $\mathcal{T}\in \mathbb{R}^{(N_h + N_f)\times C}$, where $N_h$ represents the number of historical frames and $N_f$ denotes the number of future frames.
Subsequently, we incorporate the ego pose embedding $\mathbf{e}$ and integrate it with the occupancy and the image spacetime tokens. 
Following this integration, the encoder, which comprises multiple masked SALT blocks (will be introduced in the following section), processes the multi-modal inputs concatenated along the temporal dimension. This architecture is designed to effectively model the spatial-temporal relationships among the various inputs.

\subsection{Decoder}
Following the acquisition of the encoded tokens, we develop three decoders: the occupancy decoder, the image decoder, and the pose decoder, to predict future observations. 
To enhance the learning process, we recast the task of future occupancy prediction as a voxel flow estimation problem, incorporating a warping process. 
Consequently, our occupancy decoder is composed of a flow prediction, a warping process, and a refinement module, which collectively facilitates precise and effective future occupancy predictions.

\myparagraph{Flow Prediction.}
The flow decoder, consisting of multiple stacked SALT blocks, processes encoded historical BEV features to predict future flows based on the current ego coordinate. We then convert the absolute flow map into future frames using the current-to-future transformation matrix.
These processes can be formulated as:
\begin{align}
    \mathcal{\hat{F}}_{T+1:T+{N_f}} &= \texttt{FlowDecoder}(\mathcal{Y}_h), \\
    \mathcal{\tilde{F}}_{T+1:T+{N_f}} &= R\cdot \mathcal{\hat{F}}_{T+1:T+{N_f}} + T,
\end{align}
where $R$ and $T$ are the current ego to future ego rotation and translation matrix. $\mathcal{Y}_h$ is the encoded historical occupancy features.

\myparagraph{Dynamic Decoupled Voxel Warping and Refinement.}
As illustrated in Fig. \ref{fig3:inner}(b), we propose a decoupled dynamic flow approach to simplify the occupancy forecasting problem. By leveraging the semantics of occupancy, we are able to decouple the dynamic and static grids, thereby enhancing the model's interpretability and effectiveness. For the dynamic voxels, we forecast future voxel features by applying a warping operation to the current frame's occupancy features using the predicted flow:
\begin{align}
    \mathcal{Y}_f = \texttt{grid\_sample}(\mathcal{Y}_{h}^{T}, \mathcal{\tilde{F}}).
\end{align}
For the static ones, we could easily transform them to the future by assuming that the flow is 0.
Finally, we apply several CNNs to enhance the coarse warped features.

\myparagraph{Image Decoder.}
In the case of the image decoder, we first select the corresponding global temporal embedding tokens to serve as its input. Subsequently, these decoder tokens, in conjunction with the tokens derived from the encoder, are processed through multiple SALT blocks to generate future images.

\myparagraph{Pose Decoder.}
Building on prior research~\cite{zheng2023occworld,jiang2023vad}, we initialize $N_f$ future ego queries and augment them with global temporal positional embeddings. Subsequently, self-attention and cross-attention mechanisms, utilizing the predicted future occupancy features, are employed to decode future ego poses. In this context, the interaction with the forecasted occupancy may impose constraints essential for trajectory planning.

\subsection{Image-assisted Training Strategy} 
\label{sec:training}
The flexibility of our architecture, combined with its ability to concurrently predict multiple future occupancies and images, allows for the easy imposition of various constraints among the outputs.
In this context, we propose a rendering-based photometric consistency approach to enhance the feature alignment across multiple occupancy predictions. Furthermore, the use of masked attention in SALT blocks enhances occupancy forecasting performance without incurring additional computation during inference.

\begin{table*}[t]
\centering
\footnotesize
\begin{tabular}{l|cc|cccc|cccc}
\toprule
\multirow{2}{*}{Method} & \multirow{2}{*}{Input} & \multirow{2}{*}{Aux. Sup.} & \multicolumn{4}{c|}{mIoU (\%) $\uparrow$} & \multicolumn{4}{c}{IoU (\%) $\uparrow$}\\
& & & 1s & 2s & 3s & Avg.  & 1s & 2s & 3s & Avg. \\
\midrule
OccWorld-O~\cite{zheng2023occworld} & 3D-Occ & None  & 25.78 & 15.14 & 10.51 & 17.14 & 34.63 & 25.07 & 20.18 & 26.63 \\
RenderWorld~\cite{yan2024renderworld} & 3D-Occ & None & 28.69 & 18.89 & 14.83 & 20.80 & 37.74 & 28.41 & 24.08 & 30.08 \\
OccLLaMA~\cite{wei2024occllama} & 3D-Occ & None & 25.05 & 19.49 & \textbf{15.26} & 19.93 & 34.56 & 28.53 & 24.41 & 29.17 \\
Ours-O & 3D-Occ & None  & \cellcolor{lightgray}\textbf{31.68} & \cellcolor{lightgray}\textbf{21.29} & \cellcolor{lightgray}{15.18} & \cellcolor{lightgray}\textbf{22.71}  & \cellcolor{lightgray}\textbf{40.28}
    & \cellcolor{lightgray}\textbf{31.24} & \cellcolor{lightgray}\textbf{25.29} & \cellcolor{lightgray}\textbf{32.27} \\
\midrule
OccWorld-V~\cite{zheng2023occworld} & Camera & 3D-Occ & 11.55 & 8.10 & 6.22 & 8.62  & 18.90 & 16.26 & 14.43 & 16.53\\
Ours-V & Camera & 3D-Occ & \cellcolor{lightgray}\textbf{13.38} & \cellcolor{lightgray}\textbf{10.16} & \cellcolor{lightgray}\textbf{7.96} & \cellcolor{lightgray}\textbf{10.50} & \cellcolor{lightgray}\textbf{19.18} & \cellcolor{lightgray}\textbf{16.85} & \cellcolor{lightgray}\textbf{15.02} & \cellcolor{lightgray}\textbf{17.02} \\
\bottomrule
\end{tabular}
\vspace{-0.2cm}
\caption{\textbf{4D occupancy forecasting performance on Occ3D-nuScenes \texttt{val} set.} \textit{Aux. Sup.} indicates auxiliary supervision. \textit{-O} denotes using ground-truth 3D occupancy. \textit{-V} denotes the occupancy obtained from vision inputs. The best results are marked in bold.}
\label{table:4docc}
\vspace{-0.25cm}
\end{table*}

\noindent\textbf{Volume Rendering.}
By incorporating camera intrinsic and external parameters, we are able to compute the corresponding 3D ray for each pixel in the 2D image. 
After that, we can sample $N_p$ points $\{p_i=(x_i, y_i,z_i))\}_{i=1}^{N_p}$ along the ray in pixel $(u, v)$, the rendered depth $\hat{d}$ at this pixel can be calculated via a weighted sum on the sampled points along the ray:
\begin{align} \label{eq1}
	T_i &= \mathrm{exp}(\sum\nolimits_{j=1}^{i-1}\sigma(p_j) \delta_j), \\
	\hat{d}(u, v) &=  \sum\nolimits_{i=1}^{N_p} T_i (1-\mathrm{exp}(-\sigma(p_i) \delta_i)){d}(p_i),
\end{align}
where $d(\cdot)$ and $\sigma(\cdot)$ are the distance and volume density of the sampled point, respectively.
Besides, $\delta_i = {d}(p_{i+1}) - {d}(p_i)$ is the distance between two adjacent sampled points.
As shown in Fig.~\ref{fig3:inner}(c), we obtain the corresponding density for the sampled points by gathering original volume features and feeding them into an MLP. 
Finally, we obtain depth maps in $i$-th perspective view by collecting results from all pixels, \ie, $\mathcal{D}_i = \{\hat{d}(u, v)~|~u \in [1, H], v\in[1, W]\}$, where $(H, W)$ is the size of view image. More details are provided in the supplementary material.

\noindent\textbf{Rendering-based Photometric Consistency (RPC).}
To further enhance the temporal consistency of forecasted occupancy, we introduce the rendering-based photometric consistency (RPC). 
Given the occupancy prediction of future $f$ frames, we leverage the volume rendering technique to obtain the rendered dense depth maps for each frame (target frame as in Fig.~\ref{fig3:inner}(c)).
Similar to~\cite{zhang2023occnerf}, we minimize the multi-frame photometric projection error with the help of the rendered depths.
For each target frame, we utilize the rendered depth maps to reproject the adjacent frames, \ie source images, into the target frame:
\begin{equation}
  I_{t^{\prime}\rightarrow t} = I_{t^{\prime}} \left< \mathrm{proj}(D_t,T_{t\rightarrow t^{\prime}}, K)\right>,
  \label{eq:depth_ct}
\end{equation}
where $\left< \right>$ is the grid sampling operator, $\mathrm{proj}(\cdot)$ is the resulting 2D coordinates of the projected depths $D_t$ in $I_{t^\prime}$. $T_{t\rightarrow t^{\prime}}$ and $K$ are the transformation matrix from the target frame to source frames and the camera intrinsic matrix, respectively. After that, we compute the photometric reconstruction error $pe(\cdot)$ via
\begin{equation}
  pe(I_a,I_b) = \frac{\alpha}{2}(1-\mathrm{SSIM}(I_a,I_b)) + (1-\alpha)\Vert I_a - I_b \Vert_1,
  \label{eq:depth_ct}
\end{equation}
where $\alpha$ is set to $0.85$ by default.  Followed by~\cite{cao2022monoscene}, we use L1 and SSIM to compute $pe$.
The final loss $\mathcal{L}_{RPC}$ can be formulated as
\begin{equation}
  \mathcal{L}_{RPC} = \frac{1}{N}\mathop{\mathrm{min}}\limits_{t^{\prime}}pe(I_t,I_{t^{\prime}\rightarrow t}), 
  \label{eq:depth_ct}
\end{equation}
where $N$ is the number of target frames used for loss computation. 
We also employ the per-pixel minimum projection loss and auto-masking stationary pixels techniques introduced in~\cite{cao2022monoscene}.

\begin{table*}[t]
\centering

\footnotesize
\begin{tabular}{l|lc|cccc|cccc}
\toprule
\multirow{2}{*}{Method} & \multirow{2}{*}{Input} & \multirow{2}{*}{Aux. Sup.} & \multicolumn{4}{c|}{L2 (m) ↓} & \multicolumn{4}{c}{Collision Rate (\%) ↓}\\
    &    &      & 1s   & 2s   & 3s   & \textbf{Avg.} & 1s    & 2s    & 3s    & \textbf{Avg.}      \\
\midrule
FF~\cite{hu2021safe}                      & LiDAR & Freespace & 0.55 & 1.20 & 2.54 & 1.43 & 0.06 & 0.17 & 1.07 & 0.43 \\
EO~\cite{khurana2022differentiable}                       & LiDAR & Freespace & 0.67 & 1.36 & 2.78 & 1.60 & \textbf{0.04} & {0.09} & \underline{0.88} & \underline{0.33} \\
\midrule
UniAD~\cite{hu2023planning}                & Camera & Map,Box,Motion,Tracklets,Occ & 0.48 & \underline{0.96} & \textbf{1.65} & \underline{1.03} & \underline{0.05} & \textbf{0.17} & \textbf{0.71} & \textbf{0.31} \\
VAD~\cite{jiang2023vad}               & Camera & Map,Box,Motion  & 0.54 & 1.15 & 1.98 & 1.22 & {0.10} & \underline{0.24} & 0.96 & {0.43} \\
OccNet~\cite{tong2023scene}    & Camera & 3D-Occ,Map,Box  & 1.29 & 2.13 & 2.99 & 2.14 & 0.21 & 0.59 & 1.37 & 0.72 \\
\midrule
OccNet~\cite{tong2023scene}  & 3D-Occ & Map,Box          & 1.29 & 2.31 & 2.98 & 2.25 & 0.20 & 0.56 & 1.30 & 0.69  \\
OccWorld-O~\cite{zheng2023occworld}   & 3D-Occ & None                & 0.43 & {1.08} & 1.99 & 1.17 & {0.07} & 0.38 & 1.35 & 0.60 \\
RenderWorld-O~\cite{yan2024renderworld}   & 3D-Occ & None                & \textbf{0.35} & \textbf{0.91} & 1.84 & \underline{1.03} & \underline{0.05} & 0.40 & 1.39 & 0.61 \\
OccLLaMA-O~\cite{wei2024occllama} & 3D-Occ & None & \underline{0.37} & 1.02 & 2.03 & 1.14 & \textbf{0.04} & 0.24 & 1.20 & 0.49 \\
Ours-O   & 3D-Occ & None  & \cellcolor{lightgray}{0.38} & \cellcolor{lightgray}\underline{0.96} & \cellcolor{lightgray}\underline{1.73} & \cellcolor{lightgray}\textbf{1.02} & \cellcolor{lightgray}{0.07} & \cellcolor{lightgray}0.39 & \cellcolor{lightgray}{0.90} & \cellcolor{lightgray}0.45  \\
\midrule
OccWorld-V~\cite{zheng2023occworld}    & Camera & 3D-Occ              & 0.52 & 1.27 & 2.41 & 1.40 & 0.12 & 0.40 & 2.08 & 0.87\\
RenderWorld-V~\cite{yan2024renderworld} & Camera & 3D-Occ & 0.48 & 1.30 & 2.67 & 1.48 & 0.14 & 0.55 & 2.23 & 0.97 \\
OccLLaMA-V$^\dagger$~\cite{wei2024occllama} & Camera & 3D-Occ & 0.38 & 1.07 & 2.15 & 1.20 & 0.06 & 0.39 & 1.65 & 0.70 \\
Ours-V    & Camera & 3D-Occ              & \cellcolor{lightgray}{0.42} & \cellcolor{lightgray}1.14 & \cellcolor{lightgray}2.19 & \cellcolor{lightgray}1.25 & \cellcolor{lightgray}0.09 & \cellcolor{lightgray}0.19 & \cellcolor{lightgray}1.37 & \cellcolor{lightgray}0.55 \\
\bottomrule
\end{tabular}
\vspace{-0.2cm}
\caption{\textbf{Motion planning performance on nuScenes \texttt{val} set.} \textit{Aux. Sup.} denotes auxiliary supervision apart from the ego trajectory. We use bold and underlined numbers to denote the best and second-best results, respectively. OccLLaMA-V$^\dagger$ uses a more powerful occupancy prediction results from FBOCC~\cite{li2023fb}.}
\label{table:motion_plan}
\end{table*}

\myparagraph{Masked Attention.}
As illustrated in Fig.~\ref{fig3:inner}(a), we introduce an attention mask within the transformer layers of the encoder's SALT blocks to restrict self-attention from occupancy tokens to image tokens. This modification enables the model to function effectively in the absence of image inputs. Therefore, the model can be independently deployed without image inputs.

\begin{table}[t]
\centering
\vspace{-0.1cm}
\resizebox{\columnwidth}{!}{
    \begin{tabular}{l|ccccccc} 
        \toprule
        \multicolumn{1}{c|}{\multirow{2}{*}{Method}} & \multicolumn{7}{c}{Chamfer Distance (m$^2$) $\downarrow$} \\
        \multicolumn{1}{c|}{} &  0.5s & 1.0s & 1.5s & 2.0s & 2.5s & 3.0s & Avg.\\
        \midrule
        ViDAR~\cite{yang2023vidar} & 1.34 & 1.43 & 1.51 & 1.60 & 1.71 & 1.86 & 1.58 \\
        Ours-O & \cellcolor{lightgray}\bf0.38 & \cellcolor{lightgray}\bf0.72 & \cellcolor{lightgray}\bf0.74 & \cellcolor{lightgray}\bf0.75 & \cellcolor{lightgray}\bf0.79 & \cellcolor{lightgray}\bf0.86 & \cellcolor{lightgray}\bf0.70 \\
        Ours-V & \cellcolor{lightgray}0.40 & \cellcolor{lightgray}0.75 & \cellcolor{lightgray}0.78 & \cellcolor{lightgray}0.83 & \cellcolor{lightgray}0.89 & \cellcolor{lightgray}0.90 & \cellcolor{lightgray}0.76 \\
        \bottomrule
        \end{tabular}
}
\vspace{-0.2cm}
\caption{\textbf{Point cloud forecasting performance on OpenScene mini \texttt{val} set}. Our method surpasses prior state-of-the-art methods on future point cloud prediction.}
\label{tab:pc_forecast}
\vspace{-0.55cm}
\end{table}

\subsection{Final Objective}
Our final loss function is composed of four terms with corresponding balancing weights among all future frames, \ie, the cross entropy loss and the Lovasz-softmax loss for 4D occupancy forecasting, the L2 loss for image forecasting, the L1 loss for pose regression, and the losses for image-assisted training:
\begin{equation}
  \mathcal{L} = \lambda_1\mathcal{L}_{occ} + \lambda_2\mathcal{L}_{img} + \lambda_3\mathcal{L}_{pose} + \lambda_4\mathcal{L}_{RPC}.
\end{equation}

\section{Experiments}
\label{sec:experiments}
\subsection{Experimental Setups}
\textbf{Datasets.} 
We conduct three tasks to evaluate the performances of our method on two challenging datasets: 4D occupancy forecasting on the {Occ3D-nuScenes} dataset~\cite{tian2023occ3d}, motion planning on the {nuScenes} dataset~\cite{caesar2020nuscenes} and point cloud forecasting on the {OpenScene}~\cite{openscene2023} dataset. The dataset details are presented in the supplementary material.

\noindent\textbf{Tasks and Metrics.}
1) {4D Occupancy Forecasting:} 
The 4D occupancy forecasting task requires models to forecast the future 3D occupancy given the historical occupancy inputs, which could capture the temporal evolution of the surrounding driving scenes. The mIoU and IoU are leveraged as the evaluation metric for this task.
2) {Motion Planning:} Motion planning involves finding a collision-free path to safely and efficiently navigate the vehicle from its starting point to its destination while adhering to traffic regulations and avoiding obstacles. 
Here, we use L2 error and collision rate as the evaluation metric. 
3) Point Cloud Forecasting: 
Point cloud forecasting aims to predict future point clouds from past observations, where the evaluation metric is Chamfer Distance~\cite{yang2023vidar,khurana2023point}. 

\begin{table*}[h]
\footnotesize
\centering
\begin{tabular}{l|cc|cc|ccc|cc|cc}
    \toprule
    \multirow{2}{*}{Method} & \multicolumn{2}{c|}{Modality} & \multirow{2}{*}{Flow} & \multirow{2}{*}{$D^2$ Flow}&  \multirow{2}{*}{$\mathcal{L}_{RPC}$} & \multirow{2}{*}{$\mathcal{L}_{img}$} & \multirow{2}{*}{M. SALT} & \multicolumn{2}{c|}{Forecast}  &
    \multicolumn{2}{c}{Planning} \\
    & Train & Infer & & & & & & mIoU$\uparrow$ & IoU$\uparrow$ & L2$\downarrow$ & Col.$\downarrow$  \\
    \midrule
    OccWorld~\cite{zheng2023occworld} & Occ & Occ& & &  & &  & 17.14 & 26.63 & 1.17 & 0.60 \\
    Baseline & Occ & Occ& & &  &  &  & 18.97 & 28.44 & 1.15 & 0.58 \\
    Model A & Occ & Occ& \checkmark & &  &  &  & 20.67 & 29.89 & 1.12 & 0.54\\
    Model B & Occ & Occ& \checkmark & \checkmark &  &  &  & 20.89 & 30.97 & 1.10 & 0.52 \\
    Model C & Occ + Img & Occ& \checkmark & \checkmark & \checkmark & &  & 21.96 & 31.88 & 1.12 & 0.50 \\
    Model D & Occ + Img & Occ& \checkmark & \checkmark & \checkmark & \checkmark &   & 21.25 & 31.12 & 1.10 & 0.52 \\
    Model E & Occ + Img & Occ& \checkmark & \checkmark & \checkmark & \checkmark & \checkmark & \cellcolor{lightgray}22.71 & \cellcolor{lightgray}32.27 & \cellcolor{lightgray}1.02 & \cellcolor{lightgray}0.45 \\
    \color{gray} Model F  & \color{gray}Occ + Img & \color{gray}Occ + Img & \color{gray}\checkmark & \color{gray}\checkmark & \color{gray}\checkmark & \color{gray}\checkmark & & \color{gray}23.05 & \color{gray}32.92 & \color{gray}1.00 & \color{gray}0.42 \\
    \bottomrule
\end{tabular}%
\vspace{-0.2cm}
 \caption{\textbf{Ablation studies.} We report average results over the 1s, 2s, and 3s. $D^2$ Flow: Decoupled Dynamic Flow, M. SALT: Masked SALT, Col.: Collision rate.}
\label{tab:abl_flow}
\vspace{-0.25cm}
\end{table*}

\begin{figure*}[t]
\centering
    \includegraphics[width=0.9\linewidth]{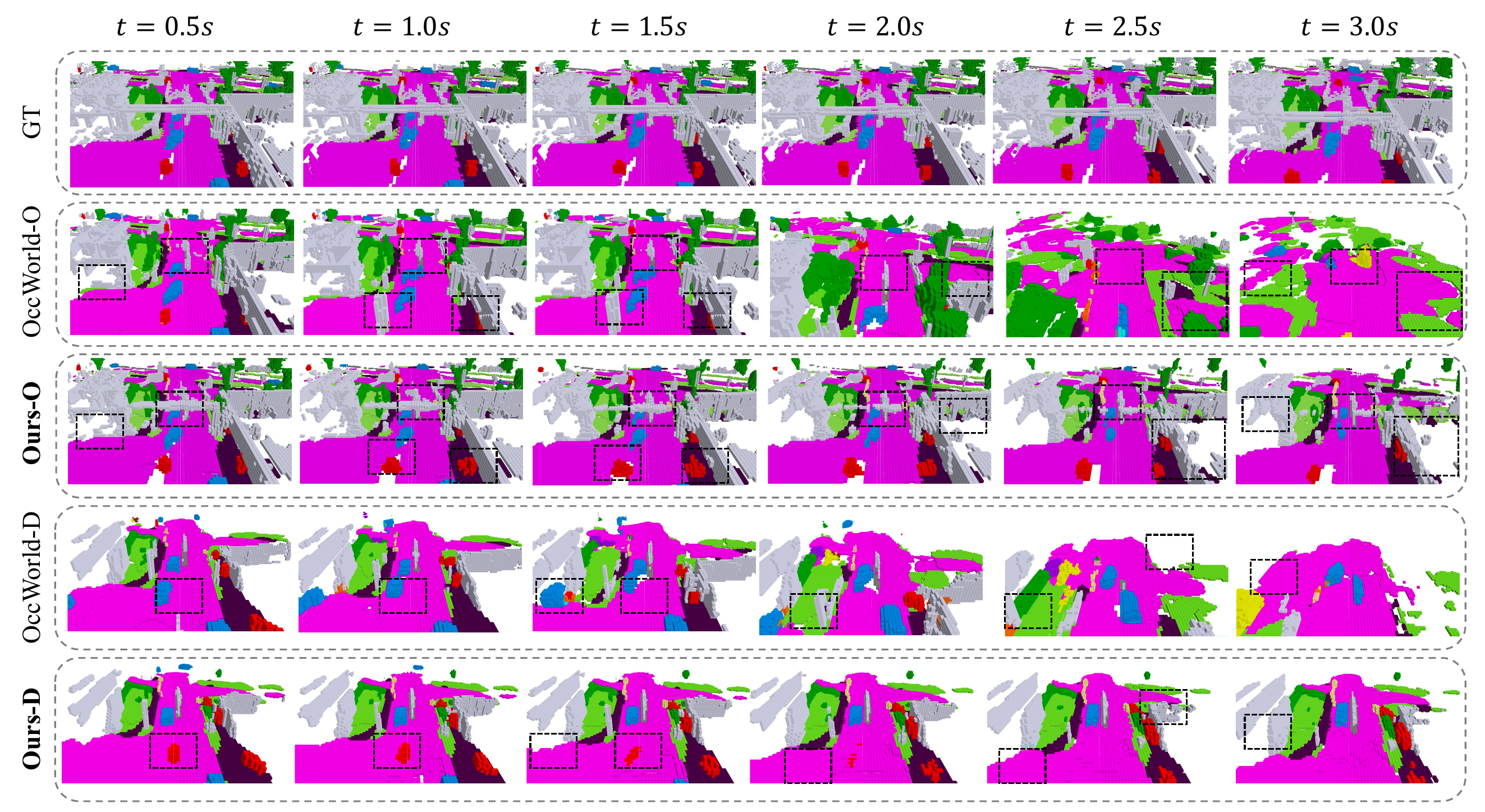}
    \vspace{-0.3cm}
    \caption{\textbf{Qualitative results of 4D occupancy forecasting.} Our method could foresee more reasonable drivable areas along with the semantic scene evolutions. However, the results generated by existing methods are inferior in the temporal consistency and dynamic object movements. Especially for the long-term predictions.}
\label{fig:quali}
\end{figure*}

\subsection{Main Results}

\noindent\textbf{4D Occupancy Forecasting.}
As presented in Tab.~\ref{table:4docc}, our method consistently demonstrates superior performance compared to OccWorld across various settings.
Specifically, when utilizing occupancy ground truth as input, our method achieves noteworthy enhancements in both mIoU and IoU, with improvements of 5.57\% and 5.64\%, respectively.
Even when employing visually approximated coarse occupancy as input, our method still maintains a stable advantage over OccWorld. This demonstrates the superiority of our method on the 4D occupancy prediction task. 
Furthermore, the qualitative visualizations provide additional support for our claims.
As depicted in Fig.~\ref{fig:quali}, our method yields cleaner, higher-quality, and more reasonable occupancy predictions for future frames. In contrast, OccWorld is prone to predicting unreasonable occupancy and lacks details. 

\noindent\textbf{Motion Planning.} Tab.~\ref{table:motion_plan} presents a quantitative comparison of our method with existing state-of-the-art approaches, including end-to-end autonomous driving methods,in the context of the motion planning task across various settings.
The results indicate that end-to-end autonomous driving solutions exhibit significant advantages in motion planning, particularly highlighted by UniAD~\cite{hu2023planning}, which is capable of generating more accurate, collision-free vehicle motion trajectories.
However, these end-to-end solutions typically necessitate substantial supervisory information, such as map data and bounding box annotations, etc.
In contrast, our approach supports planning future vehicle trajectories using only 3D occupancy supervision or even solely camera image input.
Specifically, when employing occupancy ground truth as inputs, our method surpasses both OccWorld and OccNet~\cite{tong2023scene}. Notably, even when relying exclusively on visual input, our method continues to maintain a competitive edge over the same OccWorld benchmark.
Furthermore, we observe approximately a 34\% reduction in collision rate for future 3s forecasting, indicating our method's enhanced capability in predicting long-term trajectories.

\myparagraph{Point Cloud Forecasting.} 
As illustrated in Tab.~\ref{tab:pc_forecast}, our method exhibits superior performance across all time steps in comparison to the state-of-the-art method.
Notably, even when utilizing BEVDet~\cite{huang2021bevdet} to predict the historical occupancy from visual images as inputs (Ours-V), our method continues to demonstrate significant improvements.

\subsection{Comprehensive Analysis}
\label{sec:abl}
To validate the effectiveness of our different designs, we conducted a comprehensive analysis in Tab.~\ref{tab:abl_flow}.

\myparagraph{Benefits of Decoupled Dynamic Flow.} 
Firstly, we establish our baseline model, in which future occupancy is forecasted directly in a non-autoregressive manner, but without the incorporation of flow maps or the use of an image-assisted training strategy.
In this configuration, our method demonstrates promising performance in both 4D occupancy forecasting and motion planning tasks compared to OccWorld, thereby validating the effectiveness of our framework.
Upon introducing the flow decoder, which employs the predicted flow map to warp the current occupancy features into the future for frame forecasting, we observe significant improvements in both the mIoU and IoU metrics (Model A).
Additionally, the planning metrics show slight enhancements.
When we implement the decoupled dynamic flow strategy (Model B), we observe further improvements in performance. This enhancement can be attributed to the effective utilization of occupancy sparsity, which optimizes the flow of information.

\noindent\textbf{Effects of Image-assisted Training.} 
Building upon the decoupled dynamic flow strategy, we conducted further ablation studies on several key design elements to validate the effectiveness of our image-assisted training module.
When we apply the RPC training strategy alone (Model C), we observe significant improvements in both tasks, thereby demonstrating the superiority of the proposed RPC module. Moreover, when the image modality is incorporated as an additional input during the training phase (Model D), while only forecasting occupancy during inference, we note a slight decline in performance. However, the introduction of masked SALT blocks within the cross-modal transformer encoder (Model E) leads to additional performance enhancements, indicating the effectiveness of our proposed image-assisted training strategy.
For comparative purposes, we also present the performance of Model D when utilizing multi-modal inputs (Model F), which further underscores the advantages of cross-modal interaction.

\noindent\textbf{Speed Analysis.} 
The training times for OccWorld and ViDAR are approximately \textbf{2.6} times and \textbf{2.5} times longer than ours, respectively. In contrast, our method achieves inference FPS rates that are \textbf{1.3} times and \textbf{1.8} times higher than those of OccWorld and ViDAR, respectively.
Specifically, the total training time of OccWorld is 66.7 hours, including 29.2 hours for training stage 1, and 37.5 hours for training stage 2. In contrast, our method requires only \textbf{25.1 hours} in a single-stage manner. 
During inference, the image-assisted training module can be discarded,  allowing our non-autoregressive prediction scheme to achieve \textbf{12.1} FPS, compared to 9.2 FPS for OccWorld. 

\section{Conclusion}
This paper introduces an efficient 3D occupancy world model that supports 4D occupancy forecasting, motion planning, and point cloud forecasting tasks.
Our method departs from the prevalent two-stage paradigm by employing a single-stage architecture. The proposed decoupled dynamic flow strategy and innovative image-assisted learning strategy significantly improve model performance. 
Our method achieves state-of-the-art performance on the challenging nuScenes and OpenScene benchmarks, surpassing existing world models across all tasks. 
These results demonstrate the efficacy of our method, paving the way for more robust and reliable autonomous driving.

{
    \small
    \bibliographystyle{ieeenat_fullname}
    \bibliography{main}
}

\end{document}